\documentclass[10pt,twocolumn,letterpaper]{article}

\usepackage{wacv}
\usepackage{times}
\usepackage{epsfig}
\usepackage{graphicx}
\usepackage{amsmath}
\usepackage{amssymb}

\usepackage{booktabs}
\usepackage{multirow}
\usepackage{enumitem}

\usepackage{comment}
\usepackage{paralist}
\usepackage{color}
\usepackage{subfigure}
\usepackage[belowskip=-3pt,aboveskip=9pt]{caption}

\usepackage{upgreek}

\usepackage{amssymb}
\usepackage{pifont}

\newcommand{\students}[1]{\textcolor{black}{#1}}

\def\wacvPaperID{524} %

\wacvfinalcopy %

\ifwacvfinal
\def\assignedStartPage{9876} %
\fi

\ifwacvfinal
\usepackage[breaklinks=true,bookmarks=false]{hyperref}
\else
\usepackage[pagebackref=true,breaklinks=true,colorlinks,bookmarks=false]{hyperref}
\fi

\ifwacvfinal
\else
\fi

\begin{document}

\title{Stylizing 3D Scene via Implicit Representation and HyperNetwork}

\author{Pei-Ze Chiang$^{\ast1,3}$\quad Meng-Shiun Tsai$^{\ast1,3}$\quad Hung-Yu Tseng$^2$\quad Wei-Sheng Lai$^2$\quad Wei-Chen Chiu$^{1,3}$ \\
$^1$National Yang Ming Chiao Tung University, Taiwan \quad$^2$University of California, Merced\\
$^3$MediaTek-NCTU Research Center, Taiwan
}

\twocolumn[{%
\renewcommand\twocolumn[1][]{#1}%
\vspace{-1em}
\maketitle
    \centering
    \renewcommand{\tabcolsep}{0.9pt} %
	\renewcommand{\arraystretch}{0.8} %
	\newcommand{\imagewidth}{0.32\linewidth}
    \begin{tabular}{ccc}
        \includegraphics[width=\imagewidth]{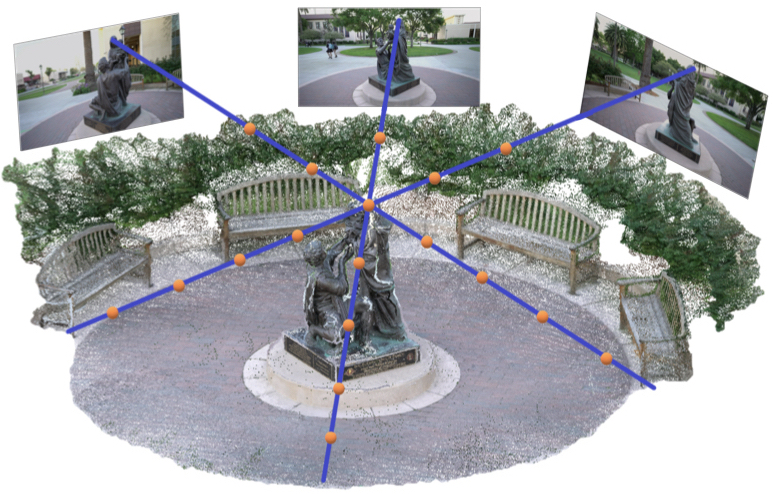} &
        \includegraphics[width=\imagewidth]{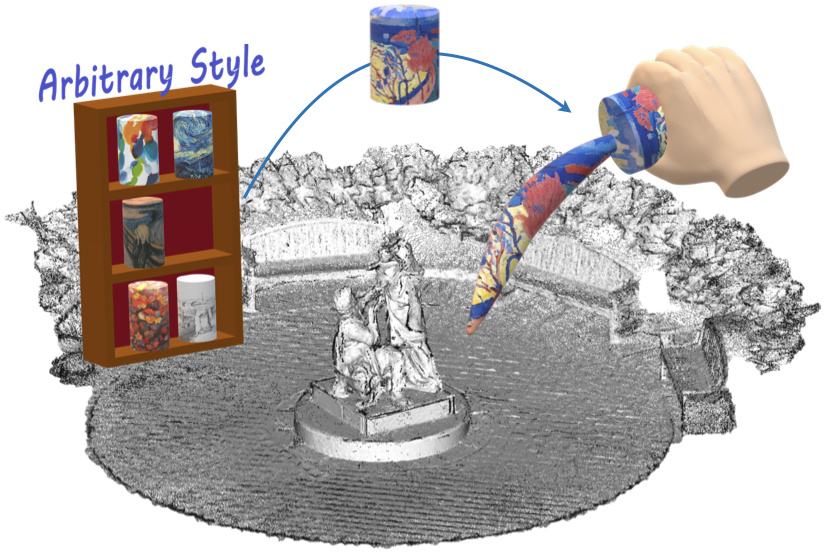} &
        \includegraphics[width=\imagewidth]{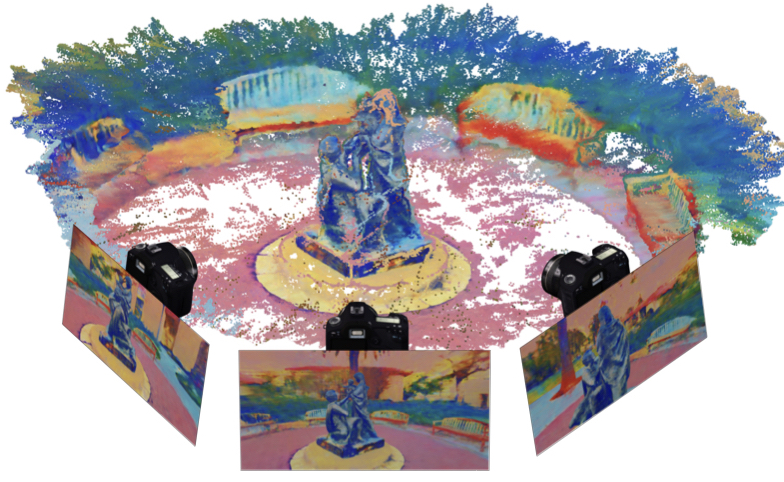} \\
        (a) Learning implicit representation &
        (b) Extracting style information &
        (c) Rendering stylized novel views \\
    \end{tabular}
    \captionsetup{hypcap=false} 
    \captionof{figure}{\textbf{Transferring arbitrary styles to complex 3D scenes.}
    \students{Our proposed 3D scene style transfer approach consists of three main parts: }
    (a) Our model first learns an implicit representation of a 3D scene that disentangles the geometry and appearance.
    (b) Then, \students{we transfer} the style information from an \emph{arbitrary} reference style image into the implicit scene representation.
    (c) Finally, our model renders stylized novel views with a consistent appearance at various view angles.\\
    }
    \label{fig:teaser}
}]
\captionsetup{hypcap=true}

\thispagestyle{empty}
\pagestyle{empty}

\begin{abstract}
\vspace{-4mm}
   In this work, we aim to address the 3D scene stylization problem - generating stylized images of the scene at arbitrary novel view angles.
   A straightforward solution is to combine existing novel view synthesis and image/video style transfer approaches, which often leads to blurry results or inconsistent appearance.
   Inspired by the high-quality results of the neural radiance fields (NeRF) method, we propose a joint framework to directly render novel views with the desired style.
   Our framework consists of two components: an implicit representation of the 3D scene with the neural radiance fields model, and a hypernetwork to transfer the style information into the scene representation.
   To alleviate the training difficulties and memory burden, we propose a two-stage training procedure and a patch sub-sampling approach to optimize the style and content losses with the neural radiance fields model.
   After optimization, our model is able to render consistent novel views at \emph{arbitrary} view angles with \emph{arbitrary} style.
   Both quantitative evaluation and human subject study have demonstrated that the proposed method generates faithful stylization results with consistent appearance across different views.
\end{abstract}
\vspace{-11mm}

\let\thefootnote\relax\footnote{$\ast$ Both authors contributed equally to the paper}

\section{Introduction}
\vspace{-1mm}
This paper focuses on the problem of \emph{stylizing complex 3D scenes}.
As shown in Figure~\ref{fig:teaser}, given a set of example images of a 3D scene and a reference image with the desired artistic style, we aim to render consistent stylized images at arbitrary novel views.
The proposed framework enables various virtual reality (VR) and augmented reality (AR) applications.
For instance, with the growing popularity of virtual tours, our method enables seamless switching between real-world scenes and virtual artistic styles, such as walking through the River Seine under Van Gogh's starry night.
3D style transfer allows us to change the style of a scene and ensures style consistency across view angles.

Numerous efforts have been made for controlling the appearance of the rendered 3D target.
For instance, Xiang~\etal~\cite{xiang2021neutex} and Kanazawa~\etal~\cite{kanazawa2018learning} formulate it as a texture synthesis problem.
Specifically, they render the 3D objects with the desired texture by aligning the coordinates of the 2D UV (texture) map to those of the target object.
However, these methods are designed specifically for a single object and are not capable of stylizing complex 3D scenes.
On the other hand, PSNet~\cite{cao2020psnet} stylizes the point cloud of a 3D scene.
Nevertheless, given a set of images of a 3D scene, it requires either the ground-truth 3D geometry or the estimated proxy geometry to build the point cloud.
Moreover, the PSNet scheme suffers from the limited resolution issue due to the discrete characteristic of the point cloud representation.
In contrast to point clouds that explicitly model 3D scenes, the recent neural radiance fields (NeRF)~\cite{mildenhall2020nerf,zhang2020nerf++,yu2020pixelnerf,schwarz2020graf} 
methods introduce an implicit representation that models a 3D scene using neural networks.
Motivated by the high-quality novel view synthesis results, we aim to leverage NeRF for transferring arbitrary styles to complex 3D scenes.

\begin{figure}[t]
    \centering
    \includegraphics[width=\columnwidth]{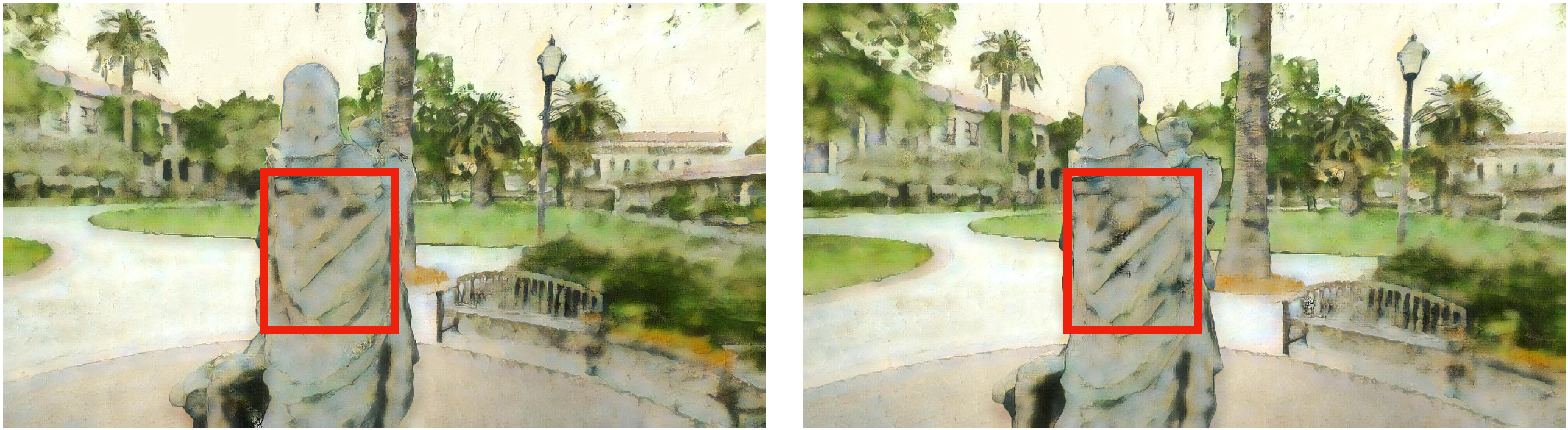}
    \caption{\textbf{Consistency issue of existing approaches.}
    Directly stylizing the complex 3D scene by combining the existing novel view synthesis (e.g.~\cite{zhang2020nerf++}) and style transfer (e.g.~\cite{huang2017arbitrary}) methods produces inconsistent results across various viewpoints for the same scene.
    Note that the red boxes highlight the inconsistent appearance in the stylized results.}
    \label{fig:c_baseline_problem}
    \vspace{-4mm}
\end{figure}

Leveraging NeRF to stylize complex 3D scenes is challenging for two reasons.
First, NeRF models lack the controllability to manipulate the appearance of the 3D scene.
Since the implicit continuous volumetric representation is built on the deep network with millions of parameters, it is unclear which parameters control the style information of the 3D scene.
To overcome this issue, one possible solution is combining existing image/video stylization approaches~\cite{huang2017arbitrary, li2017universal, li2018learning, svoboda2020two, wang2020ReReVST, deng:2020:arbitrary, gao2020fast} with novel view rendering techniques~\cite{mildenhall2020nerf,zhang2020nerf++} by first rendering novel view images and then performing image stylization.
However, as shown in Figure \ref{fig:c_baseline_problem}, current image/video stylization methods do not consider the consistency across different viewpoints for the same scene.
We empirically show that the inconsistency issue leads to various problematic results in Section~\ref{sec:exp}.
The second challenge is the memory limitation to apply the content and style losses~\cite{li2017demystifying} for learning stylization on a NeRF model.
Note that these losses are computed across \emph{holistic} images or patches in order to extract meaningful semantic features.
However, the NeRF model~\cite{zhang2020nerf++} requires dense sampling along a camera ray to render a single pixel, which takes significantly more memory to render a patch for computing the losses and back-propagate the gradients (e.g., taking $17,934$ MB to render a patch of size $67\times81$).

In this work, we propose a 3D scene style transfer approach based on NeRF to address the above mentioned challenges.
Our method is able to 1) transfer arbitrary styles to complex 3D scenes, and 2) be optimized with the commonly-used content and stylization losses.
The proposed method consists of a NeRF model and a hypernetwork.
Our NeRF model has two branches: a geometry branch and an appearance branch.
We first optimize the NeRF model to reconstruct the input 3D scene, i.e.,  learn the implicit scene representation.
Then, we fix the parameters of the geometry branch and optimize the hypernetwork to predict the parameters of the appearance branch in order to render the 3D scene with the style of the reference image.
Moreover, to alleviate the GPU memory issue, we design a patch sub-sampling algorithm to train the hypernetwork using the content and stylization loss functions.
After optimizing for a specific scene, our model is able to 1) render novel views with arbitrary and unseen styles, and 2) generate consistent stylization results across various viewpoints.
We evaluate the proposed method with quantitative metrics (e.g., measuring the consistency of stylization across different viewpoints) and subjective user studies, which demonstrate that our method performs favorably against the baseline approaches on rendering more consistent stylization results.
The main contributions of this work include:
\setdefaultleftmargin{1em}{1em}{}{}{}{} 
\begin{compactitem}
\item We propose a 3D scene style transfer approach to render novel views of a complex 3D scene with desired style.
\item We develop a hypernetwork to control the appearance-related weights of the NeRF model based on the given style image. Our model is universal and able to support arbitrary style images after optimization.
\item We demonstrate that our method can synthesize stylized images that are consistent across different view angles.
\end{compactitem}

\section{Related Work}

\subsection{Novel View Synthesis}
\label{sec:related_nvs}
Novel view synthesis aims to synthesize a target image at an arbitrary camera pose from a set of source images.
Conventional approaches~\cite{seitz2006comparison, fitzgibbon2005image, sturm1996factorization} often model a scene with \emph{explicit} 3D representations, e.g., 3D meshes~\cite{fua1995object} or 3D voxels~\cite{kutulakos2000theory} based on the multi-view geometry.
This line of work relies on a large number of source images to ensure the quality of 3D models. 
Recently, structure from motion~\cite{schonberger2016structure} and multi-view stereo~\cite{schoenberger2016mvs} techniques are also widely used to build up a 3D model.
With the rapid advance of deep learning techniques, several recent approaches learn to estimate the 3D representation of a scene, such as mesh~\cite{pontes2018image2mesh}, point cloud~\cite{lin2018learning}, and 3D voxel~\cite{henderson2019learning}.
However, these methods require supervision from the ground-truth 3D representations and are only able to reconstruct a single object.
Another group of works builds the 3D representation without ground-truth supervisions.
Image based rendering approaches~\cite{riegler2020free, riegler2020stable} integrate the image features with the 3D proxy geometry (which is often reconstructed by multi-view stereo approaches), and then warp the input images to synthesize the target view. 
Different from the explicit representations used in the above schemes, the neural radiance field approaches~\cite{mildenhall2020nerf, yu2020pixelnerf, zhang2020nerf++} encode the 3D scene information into a multi-layer perceptron (MLP), which is an \emph{implicit} 3D representation.
This method takes the 3D coordinate and camera view direction as input to directly predict the RGB values and density, which does not rely on any pre-processing to obtain the proxy geometry.
In this work, we also learn the implicit 3D representation using the neural radiance field model, but focus on transferring artistic styles to the rendered novel views.

\subsection{Image and Video Style Transfer}
Given a content image and a reference style image, style transfer methods aim to synthesize an output image which shows the style of the reference image while preserving the structure of the content image. 
As a seminal work, Gatys~\etal~\cite{gatys2016image} iteratively optimize the output image via a pre-trained model to render the desired style.
Afterwards, several methods~\cite{johnson2016perceptual, ulyanov2016texture, li2017diversified} develop feed-forward networks to significantly reduce the computational cost, but can only transfer a single or a set of pre-determined styles.
To achieve arbitrary style transfer (i.e. universal style transfer), recent methods use the adaptive instance normalization (AdaIN)~\cite{huang2017arbitrary}, whitening and coloring transform (WCT)~\cite{li2017universal}, or linear transformation (LST)~\cite{li2018learning}.
Recently, TPFR~\cite{svoboda2020two} approach disentangles an image into style and content codes, and designs a two-stage peer-regularized layer to transfer the target style into the style code of the content image.

On the other hand, applying image style transfer approaches to a video frame-by-frame often results in temporal flickering and instability, as a small perturbation in the input frame may lead to significant changes in the stylized frame.
Therefore, video style transfer methods focus on addressing the temporal consistency across the video footage. 
Existing methods introduce the optical flow to calculate temporal losses~\cite{chen2020optical, gupta2017characterizing, chen2017coherent} or align intermediate feature representations~\cite{gao2018reconet, huang2017real} in order to stabilize the model prediction across nearby video frames.
Recent efforts further achieve consistent and real-time video style transfer through temporal regularization~\cite{wang2020compound, wang2020ReReVST}, multi-channel correlation~\cite{deng:2020:arbitrary}, and bilateral learning~\cite{xia2021real}.
Although these methods have demonstrated impressive performance, they are designed specifically for stylizing images or video sequences.
Since the consistency across various viewpoints of the same scene is not considered, directly using existing image/video stylization schemes for our problem has various issues (see Figure~\ref{fig:c_baseline_problem} and Section~\ref{sec:exp}).

\subsection{Texture Transfer}
Texture transfer aims to change the texture or style of a 3D object while keeping the appearance consistent across different view angles.
With the recent advance of deep learning techniques, different methods~\cite{kanazawa2018learning, groueix2018papier, xiang2021neutex} are proposed to learn the correspondences between 3D shape (e.g. meshes) and the texture space for realizing the texture transfer. 
For instance, Xiang~\etal~\cite{xiang2021neutex} learn a mapping from the implicit 3D representation to the 2D texture map, such that one can change the appearance of a 3D model by swapping the 2D texture map.
However, such a texture mapping approach may generate unnatural results if the texture image is mapped across object edges or boundaries.
In addition to texture mapping, recently there are several works~\cite{kato2018neural,cao2020psnet} proposed to tackle the style transfer task on the 3D representations. For instance,
Kato~\etal~\cite{kato2018neural} propose to build a neural renderer where the rendering is integrated into neural networks, and demonstrate its application of style transfer on 3D models. PSNet~\cite{cao2020psnet} performs the style transfer on the point cloud data via manipulating the point cloud features in latent space to change the style. However, these methods rely on explicit 3D representations, such as mesh and point cloud, and are limited to the object level instead of the entire scene. Moreover, they do not support synthesizing stylized images with high-quality, which further limits their applications in the real world (e.g. AR). In this work, we resort to the \emph{implicit} 3D scene representation and focus on transferring style for real-world 3D scenes with a complex background to generate the stylized novel views.
\section{Proposed Method}

\begin{figure*}[ht]
    \centering
    \includegraphics[width=0.97\textwidth]{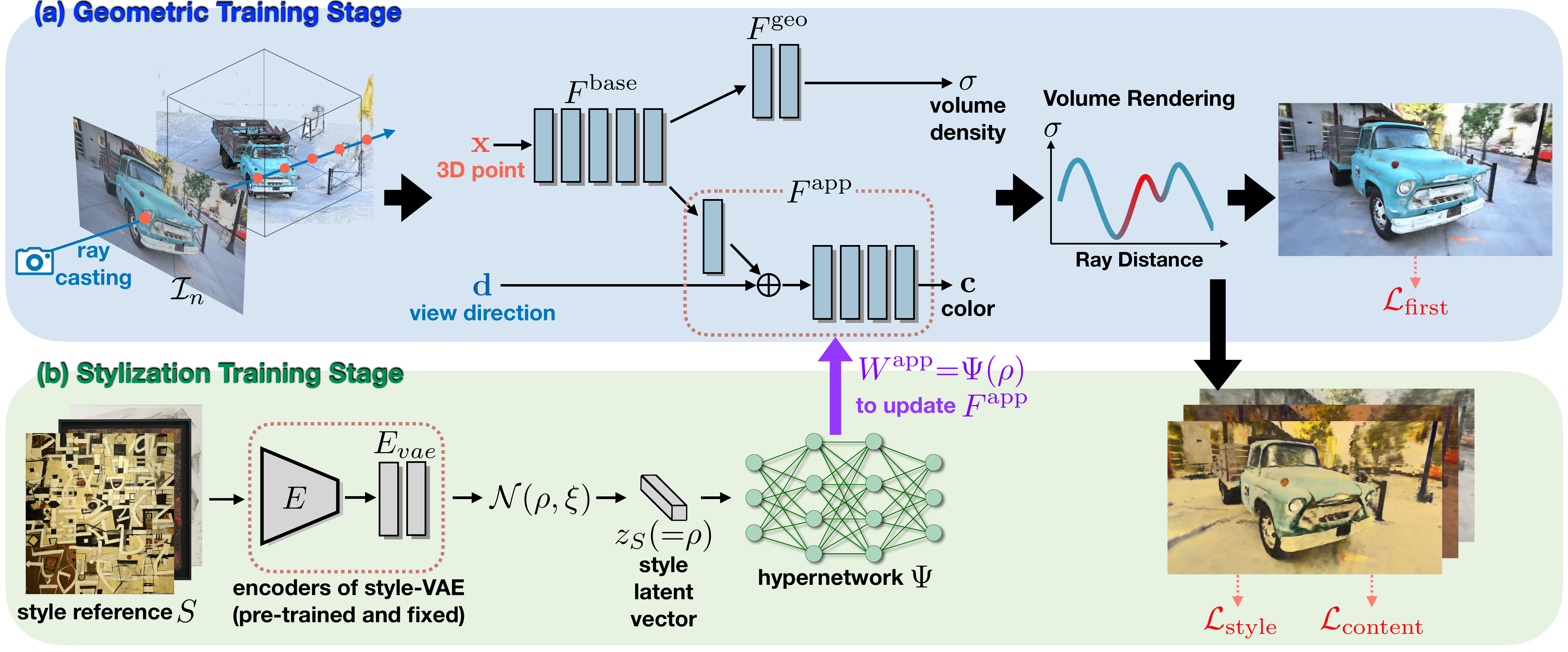}
    \caption{\textbf{Algorithmic overview.}
    The proposed framework consists of a NeRF model with a geometry branch (i.e. $F^{\text{base}}$ together with $F^{\text{geo}}$) as well as an appearance branch (i.e. $F^{\text{app}}$), and a hypernetwork $\Psi$.
    (a) Firstly in the geometric training stage, it learns to reconstruct the target 3D scene from a set of input images $\{\mathcal{I}_n\}_{n=1}^N$.
    (b) Then in the stylization training stage, it learns to transfer arbitrary style. We keep the geometry branch fixed, and optimize the hypernetwork $\Psi$ to predict the parameters $W^{\text{app}}$ in the appearance branch according to the style latent vector $z_S$ extracted from the input reference style image $S$.
    After two-stage optimization, we use volume rendering to render images with desired style at arbitrary novel views.
    }
    \label{fig:b_model_archi}
    \vspace{-2mm}
\end{figure*}

Given a set of $N$ images/photos $\{\mathcal{I}_n\}_{n=1}^N$ of a static 3D scene taken from different camera poses $\{ {(R_n, t_n)} \}_{n=1}^N$, our goal is to render arbitrary novel views of the 3D scene with the style extracted from a reference image $S$.
The rendered images should have consistent appearance and stylization effect across different views.
To this end, we propose a 3D style transfer method to enable the universal stylization of a complex 3D scene.
We model a 3D scene with implicit representation by the neural radiance fields (Section~\ref{sec:nerf++}), and learn to transfer arbitrary style using a hypernetwork (Section~\ref{sec:hyper}).
To alleviate the training difficulties, we propose a two-stage training pipeline, 
where the geometric training stage learns the implicit representation of a 3D scene by disentangling the geometry and appearance into two branches, 
and the stylization training stage learns to predict the parameters of the appearance branch from the reference image $S$ (Section~\ref{sec:model_training}).

\subsection{Preliminaries}
\label{sec:nerf++}
The model of neural radiance fields (NeRF) proposed in~\cite{mildenhall2020nerf} adopts a sparse set of input views of a 3D scene for learning to optimize the underlying continuous volumetric scene function.
The basic idea behind NeRF can be illustrated in Figure~\ref{fig:b_model_archi}(a). Given a camera observing the 3D scene at the viewing direction $\mathbf{d}=(\theta,\phi)$, we first march along the rays back-projected from the camera center through all the pixels on the image plane for obtaining the samples of 3D points $\mathbf{x}=(x, y, z)$. The scene function as an implicit scene representation then takes both $\mathbf{x}$ and $\mathbf{d}$ as input to output the volume density $\sigma$ at $\mathbf{x}$ and the corresponding RGB color $\mathbf{c}=(r, g, b)$ emitted towards the viewing direction $\mathbf{d}$. To be detailed, the scene function consists of three multilayer perceptrons (MLPs): $F^{\text{base}}$, $F^{\text{geo}}$, and $F^{\text{app}}$. In practice, $F^{\text{base}}$ takes $\mathbf{x}$ as input where the produced $F^{\text{base}}(\mathbf{x})$ is either passed through $F^{\text{geo}}$ to obtain the volume density $\sigma$, or further processed by $F^{\text{app}}$ together with $\mathbf{d}$ to obtain the view-dependent color $\mathbf{c}$ via $F^{\text{app}}(F^{\text{base}}(\mathbf{x}), \mathbf{d})$. With accumulating the colors and densities via the volume rendering techniques, the high-quality 2D images as the observation of the 3D scene from various views can be generated. Please note that in practical implementation~\cite{mildenhall2020nerf} both $\mathbf{x}$ and $\mathbf{d}$ are firstly transformed into positional embeddings before being utilized by the MLPs (i.e. $F^{\text{base}}$ and $F^{\text{app}}$).

While the original NeRF model seems to demonstrate compelling capability in the view synthesis, 
when it is adopted to tackle the $360^\circ$ captures of unbounded and complex scenes, simultaneously modelling the nearby and far objects (related to foreground and background respectively) by the same volumetric scene function would cause problems for volume rendering as being required to handle the large dynamic depth range between objects, as pointed out by~\cite{zhang2020nerf++}. 
To deal with such issue, NeRF++~\cite{zhang2020nerf++} separates the foreground and background objects into the inner volume and outer volume by a unit sphere, with having two NeRFs adopted to model them respectively, where an inverted sphere parameterization is particularly applied on the coordinate system of outer volume for bounding the unlimited distance between the background objects and the origin. In our work, we hence adopt the model of NeRF++~\cite{zhang2020nerf++} to represent the unbounded and complex 3D scenes.

\subsection{\students{Stylizing Implicit Representations}}
\label{sec:hyper}
Without loss of generality, the operation of style transfer aims to keep the geometry/content of the target scene while modifying its appearance/texture according to the reference style image. 
From the design of NeRF++ models, we can see that the geometry and the appearance information of the scene are respectively represented by the volume densities and the view-dependent color values at each 3D location. Therefore, to stylize the 3D scene implicitly encoded by NeRF++, we propose a novel approach to modify the parameters of the MLP (i.e. $F^{\text{app}}$, the \textbf{appearance branch}) which is responsible for predicting the color values, while keeping the other MLPs (i.e. $F^{\text{base}}$ and $F^{\text{geo}}$, the \textbf{geometry branch}) fixed to retain the geometry of the target scene. Basically, our stylization method on the NeRF-based scene representation is realized by two components: the \textbf{style variational autoencoder} (i.e., \textbf{style-VAE}) to extract the style latent vector $z_S$ from the style image $S$, and the \textbf{hypernetwork $\Psi$}~\cite{ha2016hypernetworks} to estimate the weights for modifying $F^{\text{app}}$ according to $z_S$ (as illustrated in Figure~\ref{fig:b_model_archi}). \students{We detail these two components in the supplementary material}.

\subsection{Model Training}
\label{sec:model_training}

The training procedure for our 3D scene style transfer approach contains the geometric training stage and the stylization training stage. We describe these two stages and the corresponding objective functions in the following.

\vspace{-3mm}
\paragraph{Geometric Training Stage (First Stage).}
In this stage we aim to learn a NeRF-based representation of the target 3D scene from the given $N$ images $\{\mathcal{I}_n\}_{n=1}^N$ taken from different camera poses $\{(R_n, t_n)\}_{n=1}^N$. The NeRF++~\cite{zhang2020nerf++} model is adopted and its training process is briefly summarized as follows. At each optimization iteration, we randomly sample $M$ pixels from input images to form a batch of camera rays $\{\mathbf{r}_m\}_{m=1}^M$ and 3D points (marching along the rays) based on the corresponding camera poses and camera intrinsic. With taking these 3D points and their corresponding view directions (i.e. the directions of the corresponding camera rays) as input to the scene function of NeRF++ to obtain the output set of volume densities and colors, the volume rendering technique is used to render the color value $\hat{\mathbf{c}}(\mathbf{r}_m)$ of each ray $\mathbf{r}_m$. The objective for training the MLPs of the scene function (i.e. $F^{\text{base}}$, $F^{\text{geo}}$, and $F^{\text{app}}$) is the mean square error (MSE) between $\hat{\mathbf{c}}(\mathbf{r}_m)$ and the groundtruth color $\mathbf{c}(\mathbf{r}_m)$ of the corresponding image pixel of ray $\mathbf{r}_m$.
\vspace{-2mm}
\begin{equation}
\label{eq:first_stage_loss}
\mathcal{L}_{\text{first}} = \frac{1}{M} \sum_{m=1}^M \left \| \hat{\mathbf{c}}(\mathbf{r}_m) - \mathbf{c}(\mathbf{r}_m)  \right \|_2 
\end{equation}

\vspace{-6mm}
\paragraph{Stylization Training Stage (Second Stage).}
As in the previous geometric training stage, we have encoded the complete geometry and the original appearance of the target 3D scene into the NeRF++ model, now in the stylization training stage we focus on learning the hypernetwork $\Psi$ to predict from the style latent vector $z_S$ the weights $W^{\text{app}}$ for updating MLP $F^{\text{app}}$, where $z_S$ is extracted from the style reference image $S$ by the pre-trained encoders of style-VAE, 
i.e. $E_{vae}(E(S))$. 
Note that as the geometry of the 3D scene should be retained during the stylization, both $F^{\text{base}}$ and $F^{\text{geo}}$ are kept fixed during this training stage.

\begin{figure}[t]
    \centering
    \includegraphics[width=\columnwidth]{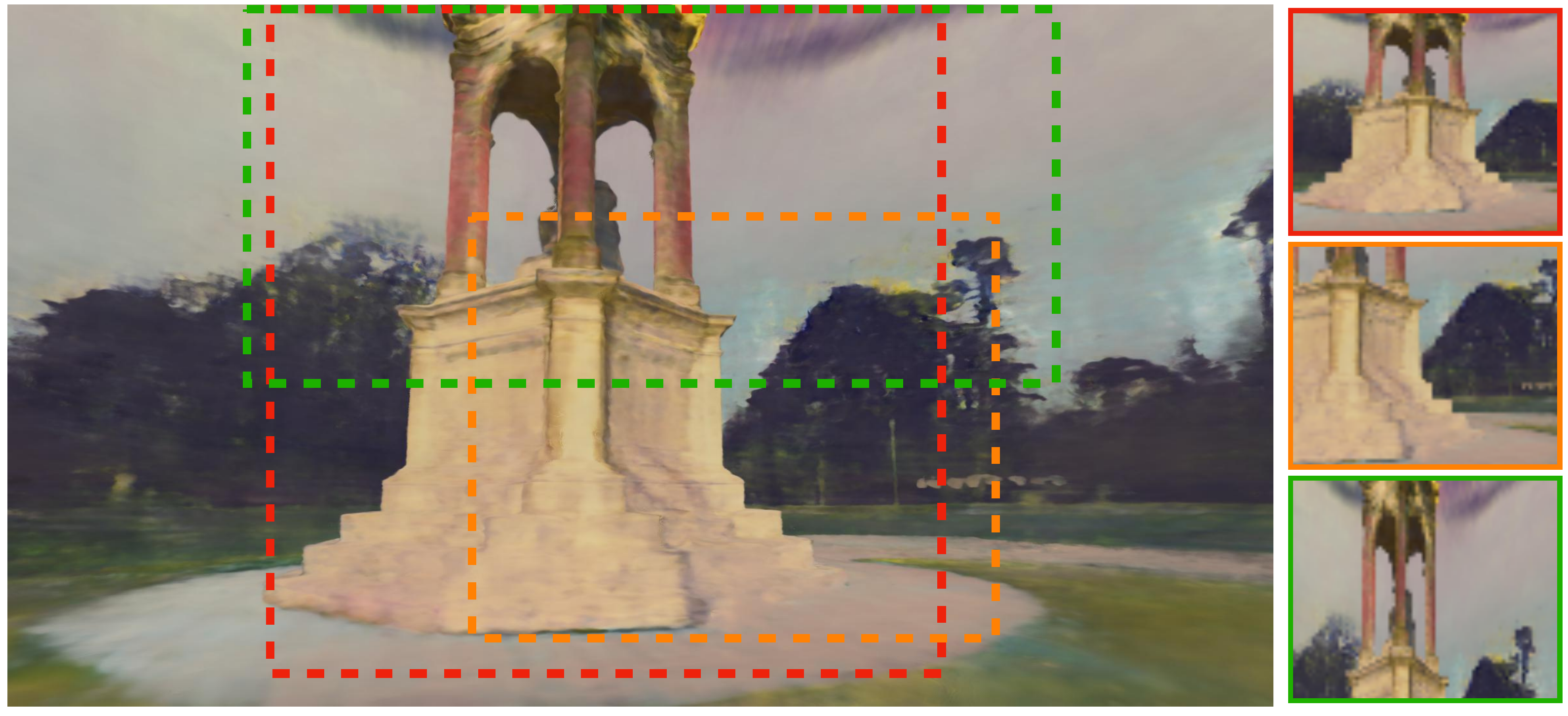}
    \caption{\textbf{Patch sub-sampling.}
    \students{We randomly crop the input image with various window sizes (i.e., dash-lined boxes) and use the nearest neighbor sub-sampling to get the cropped patch, as the three patches shown in the right.}
    }
    \label{fig:i_patch}
    \vspace{-3mm}
\end{figure}

As the goal of our style transfer is to stylize the whole scene such that the generated images of view synthesis are able to demonstrate the similar style as the reference image $S$, ideally we should render the holistic images or patches to compute typical content and style losses~\cite{huang2017arbitrary}, which are widely used in style transfer works, for validating the effectiveness of stylization. However, since rendering a single pixel already needs dense sampling on the camera ray to obtain numerous 3D points for querying the NeRF++ model, it would be even more costly in terms of GPU memory usage when attempting to render the holistic image/patch and perform the back-propagation for network optimization. In order to tackle such issue, we propose a \textbf{patch sub-sampling} algorithm to enable the computation of content and style losses.
\students{Note that Schwarz~\etal~\cite{schwarz2020graf} demonstrate that using sparse sampling rather than down-sampling can retain the high-frequency details of the real image.}
As illustrated in Figure~\ref{fig:i_patch}, a random window is firstly cropped from the input image with having the height and width of such window at least larger than a ratio ($\Lambda_w$ and $\Lambda_h$ respectively) of the input image size, then we uniformly sample $\eta_w \times \eta_h$ pixels from such window to form a smaller patch (where the number of pixels is constrained by the GPU memory). Using our patch sub-sampling algorithm to select the patch $P(\mathcal{I}_n)$ from input image $\mathcal{I}_n$ and its paired patch $P(\tilde{\mathcal{I}}_n)$ from the stylization result $\tilde{\mathcal{I}}_n$, the content and style losses are computed between them. 
In detail, the content loss $\mathcal{L}_{\text{content}}$ encourages $P(\mathcal{I}_n)$ and $P(\tilde{\mathcal{I}}_n)$ to have similar content features:
\begin{equation}
\label{eq:content_loss}
    \mathcal{L}_{\text{content}} = \left \| \tau(P(\mathcal{I}_n)) - \tau(P(\tilde{\mathcal{I}}_n)) \right \|_2
\end{equation}
where $\tau(\cdot)$ denotes the feature representation obtained from the \texttt{relu4\_1} layer of an ImageNet-pretrained VGG-19 network.
And the style loss $\mathcal{L}_{\text{style}}$ measures the mean squared error in terms of feature statistics between $P(\tilde{\mathcal{I}}_n)$ and the reference style image $S$, which is defined as:
\begin{equation}
\begin{aligned}
\label{eq:style_loss}
    \mathcal{L}_{\text{style}} &= \sum_{l} \left \| \mu(\tau_l(S)) - \mu(\tau_l(P(\tilde{\mathcal{I}}_n)))\right \|_2 \\&+ \sum_{l} \left \| \Sigma(\tau_l(S)) - \Sigma(\tau_l(P(\tilde{\mathcal{I}}_n)))\right \|_2
\end{aligned}
\end{equation}
where $\mu$ and $\Sigma$ denote the mean and standard deviation respectively, and  $\tau_l(\cdot)$ denotes the feature representation obtained from the $l$-th layer of an ImageNet-pretrained VGG-19 network, basically \texttt{relu1\_1}, \texttt{relu2\_1}, \texttt{relu3\_1}, and \texttt{relu4\_1} layers are used.

The overall objective function for the stylization training stage, in which the gradients are back-propagated to learn the hypernetwork $\Psi$, is then defined as:
\begin{equation}
\label{eq:second_stage_loss}
\mathcal{L}_{\text{second}} = \mathcal{L}_{\text{content}} + \lambda_{\text{style}}\mathcal{L}_{\text{style}}
\end{equation}
where the hyperparameter $\lambda_{\text{style}}$ controls the balance between the content loss and style loss, and we set $\lambda_{\text{style}} = 15$ for all our experiments.

\vspace{-2mm}
\paragraph{Implementation Details.}
\students{In the geometric training stage, our NeRF model is trained for $250,000$ iterations and we set $M=67\times81$; while in the stylization training stage, hypernetwork $\Psi$ is trained for $100,000$ iterations and we set  $\Lambda_w, \Lambda_h, \eta_w,$ and $\eta_h$ to $1/3, 1/2, 81,$ and $67$. We adopt the Adam optimizer for both stages with learning rates set to $0.0005$ and $0.001$, respectively. Following~\cite{shen2018neural}, each style image used in our experiments is resized and randomly cropped to be the size of $256 \times 256$.}

\section{Experimental Results and Analysis}
\label{sec:exp}

In this section, we present qualitative and quantitative results to validate the effectiveness of our proposed framework.
Please refer to 
our project page$^1$\footnote{$^1$Project page: \href{https://ztex08010518.github.io/3dstyletransfer/}{https://ztex08010518.github.io/3dstyletransfer/}}
for source code, pre-trained model and more qualitative results.

\vspace{-4mm}
\paragraph{Datasets.}
We conduct the experiments using five real-world 3D scenes collected in the Tanks and Temples~\cite{Knapitsch2017} dataset, i.e., Family, Francis, Horse, Playground and Truck.
During the geometric training stage, we follow~\cite{mildenhall2020nerf} to use the COLMAP SfM~\cite{schonberger2016structure} method to estimate the camera poses and intrinsics of the input images for each 3D scene.
On the other hand, we use $81442$ images in the WikiArt dataset~\cite{nichol2016painter} as the reference style images.
Specifically, we randomly select $112$ images as the testing data and keep the others (i.e., totally $81330$) for the stylization training stage.

\vspace{-4mm}
\paragraph{Compared Methods.}
To the best of our knowledge, there is no existing method that focuses on stylizing complex 3D scenes.
Therefore, we combine different image/video stylization methods with the novel view synthesis (NVS) algorithms to build three types of the baseline approaches:
\begin{compactitem}
\item Image stylization $\rightarrow$ NVS: we stylize the input images of the target scene, then perform novel view synthesis.
\item NVS $\rightarrow$ image stylization: we perform image stylization on the novel view synthesis results.
\item NVS $\rightarrow$ video stylization: we treat a series of novel view synthesis results (generally along a smooth camera path) as the \emph{video}, then perform video stylization.
\end{compactitem}
Specifically, we use NeRF model described in Figure~\ref{fig:b_model_archi} (a) as the novel view synthesis approach.
The AdaIN~\cite{huang2017arbitrary}, WCT~\cite{li2017universal}, LST~\cite{li2018learning}, and TPFR~\cite{svoboda2020two} schemes are used for image stylization.
Finally, we use two video stylization frameworks, i.e., ReReVST~\cite{wang2020ReReVST} and MCCNet~\cite{deng:2020:arbitrary}. 

\subsection{Qualitative Results}
\label{sec:qualitative}
\begin{figure}[t]
    \centering
    \includegraphics[width=0.95\columnwidth]{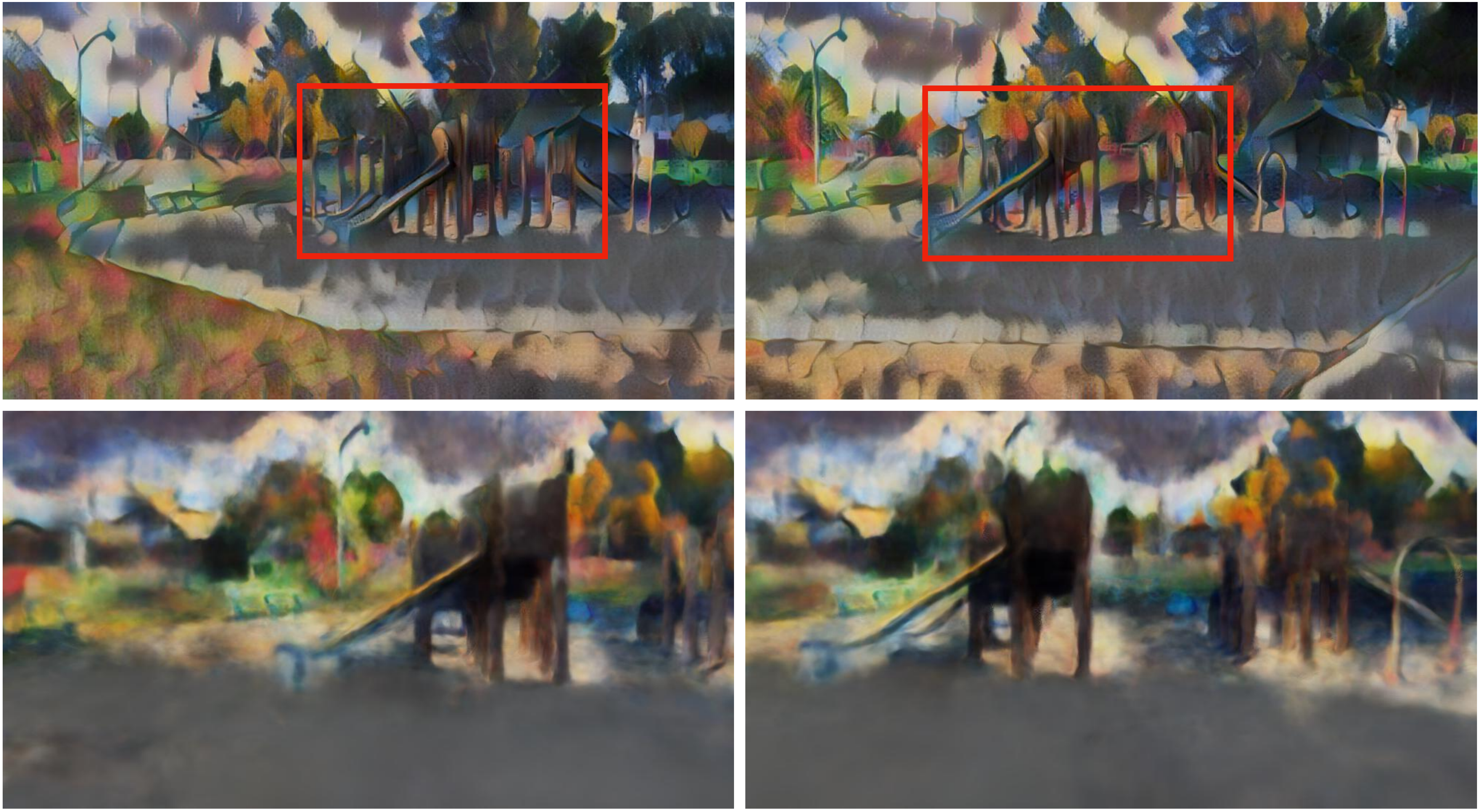}
    \caption{\textbf{Qualitative results of ``image stylization $\rightarrow$ NVS" baseline.}
    \students{
    Since the training images (top row) are stylized independently by AdaIN~\cite{huang2017arbitrary}, the stylized training images are not consistent (red boxes).
    Therefore, the novel view synthesis algorithm blends such inconsistency and produces blurry results (bottom row).}
    }
    \label{fig:visual_image2novel}
    \vspace{-5mm}
\end{figure}

\begin{figure*}[h!]
    \centering
    \includegraphics[width=.98\textwidth]{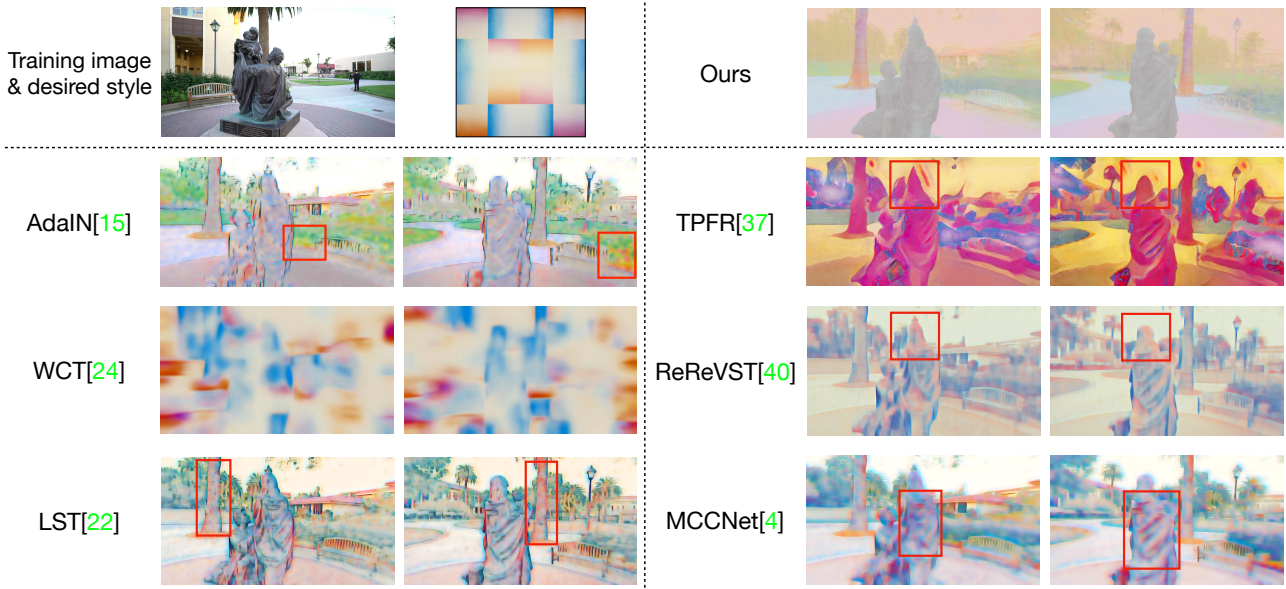}. 
    \vspace{-.8em}
    \caption{\textbf{Qualitative comparisons.}
    The bottom row presents one of the training images of the target scene with the input reference (style) image and the stylization results of our proposed approach.
    The red boxes highlight the inconsistent stylization across different views, while our proposed method is consistent across different view angles with desired style.
    }
    \label{fig:visual_baseline}
\end{figure*}

We present the qualitative comparisons in Figure~\ref{fig:visual_image2novel} and \ref{fig:visual_baseline}.
The baseline ``image stylization $\rightarrow$ NVS" produces blurry results, as shown in Figure~\ref{fig:visual_image2novel}.
Since the input images are processed independently by the AdaIN~\cite{huang2017arbitrary} approach, the stylized images are not consistent across different views of the same scene.
Therefore, the optimization of the NeRF model with these inconsistent images leads to blurry results.
Moreover, as the NeRF model is optimized for a specific style, this baseline method is not capable of transferring arbitrary style to the 3D scene.
On the other hand, the baseline ``NVS $\rightarrow$ image stylization" also produces inconsistent results across different viewpoints, as highlighted in the red boxes in Figure~\ref{fig:visual_baseline}.
In particular, the baseline based on the WCT~\cite{li2017universal} approach fails to preserve the content of the original 3D scene, while the other one based on the TPFR~\cite{svoboda2020two} scheme does not transfer the desired style provided by the reference image.
In contrast, the results synthesized by our method not only match the desired style, but also are consistent across various novel views. 

\begin{figure*}[ht!]
    \centering
    \includegraphics[width=.98\textwidth]{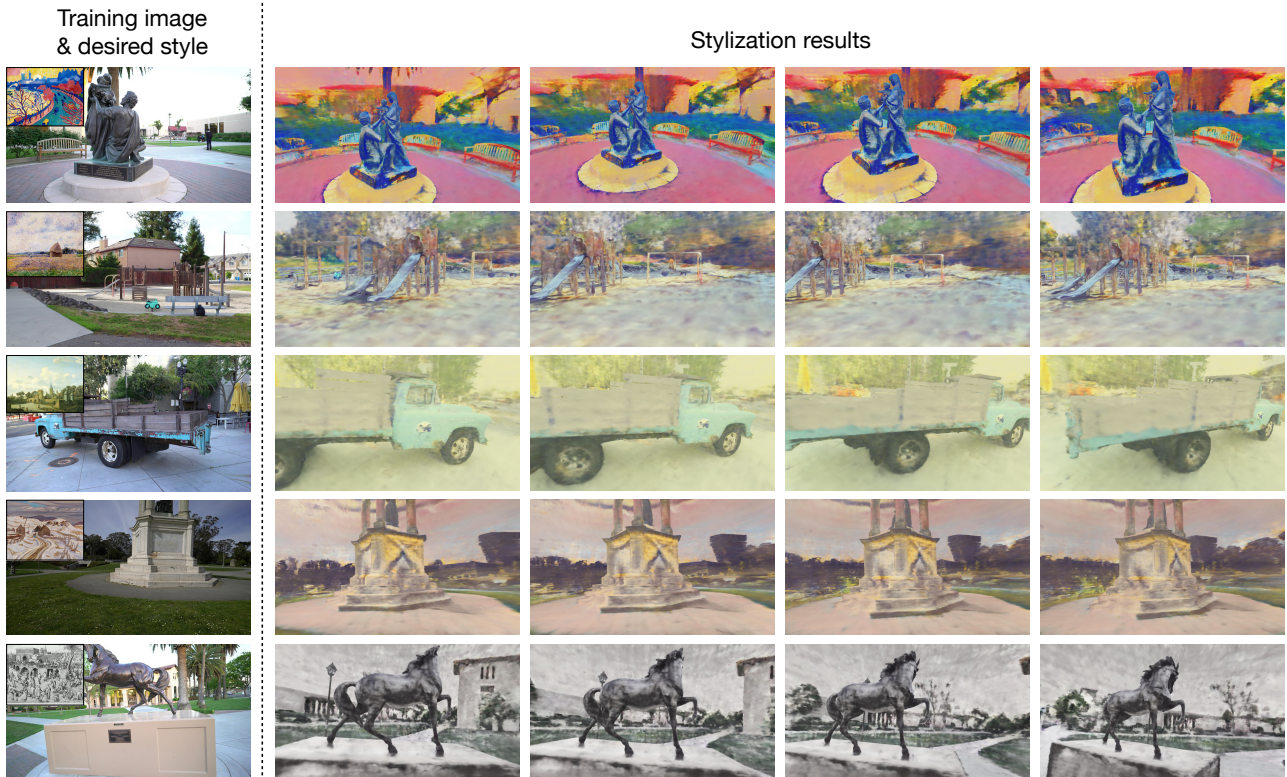}
    \vspace{-.8em}
    \caption{\textbf{Qualitative results of our proposed framework of 3D scene stylization.}
    For each row, the leftmost column presents one of the training images of the target scene together with the input reference (style) image on the top-left corner, while the remaining columns demonstrate the stylization results at various novel views.}
    \label{fig:visual_ours}
\end{figure*}

We demonstrate the qualitative results by the baseline ``NVS $\rightarrow$ video stylization" in Figure~\ref{fig:visual_baseline}.
Although these video-stylization-based methods are trained to consider the \emph{short-term} consistency, they fail to produce consistent results between two far-away viewpoints due to the error accumulation, e.g., the head and the back of the statue.
In contrast, since our framework is trained to stylize the holistic 3D scene, it generates results that are consistent between both short-range or long-range viewpoints. More stylized results of our proposed method are provided in Figure~\ref{fig:visual_ours}.

\subsection{Quantitative Results}
\label{sec:quantitative}
\paragraph{User Preference Study.}
To evaluate the quality of stylizing complex 3D scenes, we conduct a study to understand the user preference between the results rendered by the proposed and baseline methods. Here we focus on the comparison against ``NVS $\rightarrow$ image stylization" and ``NVS $\rightarrow$ video stylization" baselines as the results of ``image stylization $\rightarrow$ NVS" ones are generally blurry as shown in Figure~\ref{fig:visual_image2novel}.

\begin{figure}[!h]
    \centering
    \includegraphics[width=0.7\columnwidth]{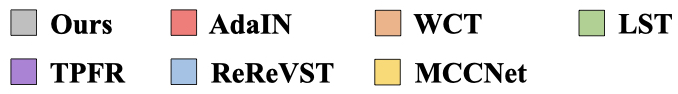} \\
    \includegraphics[width=\columnwidth]{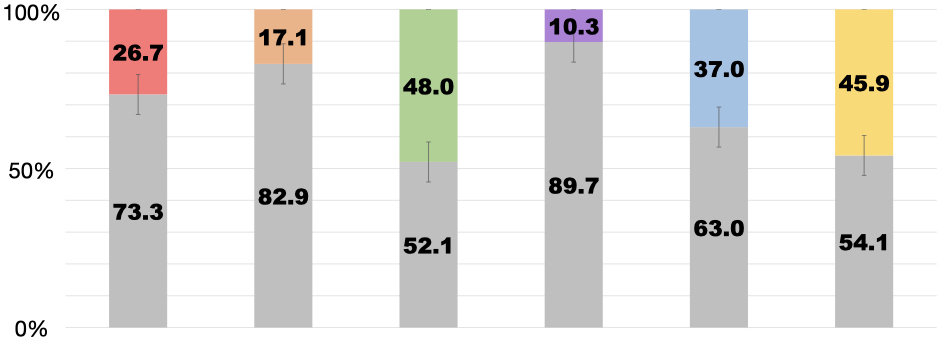} \\
    (a) Stylization \\
    \includegraphics[width=\columnwidth]{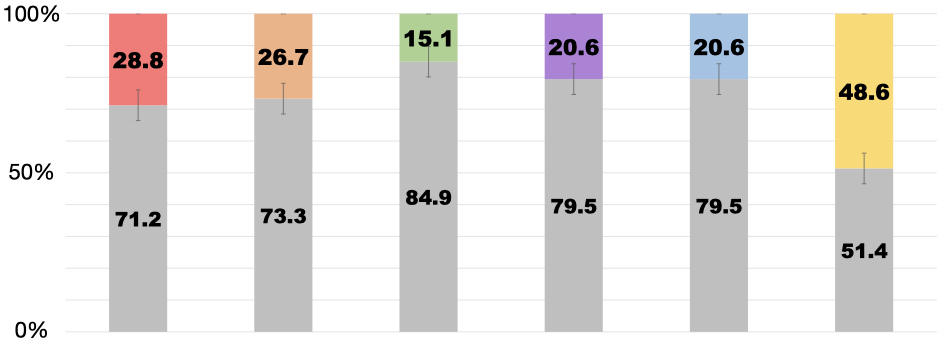} \\
    (b) Consistency \\
    \vspace{-1mm}
    \caption{\textbf{User preference study.}
    We present two novel view synthesis results (created as videos) generated by different methods, and ask the user to select the one that 1) matches the given style image and 2) shows less flickering.}
    \label{fig:user_study}
    \vspace{-5mm}
\end{figure}

There are $73$ users participated in this study.
For each user, there are $2$ tests conducted for each comparison (i.e. proposed method versus one baseline in terms of stylization quality or temporal consistency). 
As shown in Figure~\ref{fig:user_study}, our proposed method performs favorably against the baseline schemes in terms of both the stylization quality and consistency.
We also observe that the ``NVS $\rightarrow$ video stylization" baseline produces videos with less flickering compared to the ``NVS $\rightarrow$ image stylization" ones since they consider the temporal consistency.
However, these approaches fail to preserve the consistency between two far-away viewpoints, as demonstrated in the following experiments.

\vspace{-4mm}
\paragraph{Consistency.}
In addition to the user preference study that evaluates the quality of the stylization results, we use the metric from Lai~\etal~\cite{lai2018learning} to measure the consistency between different stylized novel view images. 
\students{More details of the consistency metric are provided in the supplementary materials.}
In the following experiments, we evaluate the consistency from two different perspectives: 1) the \emph{short-range} consistency between nearby novel views, and 2) the \emph{long-range} consistency between far-away novel views.

Table~\ref{tab:quant_short_consis} shows the short-range consistency scores.
In this experiment, we use every two adjacent novel views, i.e., the $t-th$ and $(t-1)-th$ frames in the testing videos, to compute the consistency score.
We observe that the results generated by the image stylization baseline methods are not consistent as the novel view images are processed independently.
Moreover, while the TPFR~\cite{svoboda2020two} approach achieves the lowest scores among all the baseline schemes, it fails to capture the desired style of the reference image in some cases, as shown in Figure~\ref{fig:visual_baseline}.

\begin{table}[!h]
\centering
\renewcommand{\arraystretch}{1.1}
\setlength{\tabcolsep}{5pt}
\vspace{-1mm}
\caption{\textbf{Quantitative comparisons on short-range consistency.}
We compute the consistency score (the lower the better) between stylized images at two adjacent novel views.
}
\resizebox{\columnwidth}{!}{
\begin{tabular}{l|ccccc|c}
\toprule
Methods         & Family            & Francis           & Horse             & Playground        & Truck             & Average           \\ \midrule
AdaIN           & 2.0172            & 1.9015            & 2.5102            & 1.7011            & 1.8582            & 1.9976            \\
WCT             & 2.9717            & 2.8170            & 3.5992            & 2.5998            & 3.0162            & 3.0008            \\
LST             & 4.3897            & 3.2608            & 4.0586            & 3.1262            & 3.6980            & 3.7067            \\
TPFR            & 1.0930            & 0.6611            & 1.2504            & 0.6448            & 0.6908            & 0.8680            \\ \midrule
ReReVST         & 1.0089            & 0.8431            & 1.3006            & 0.6404            & 0.8617            & 0.9309            \\ 

MCCNet          & 1.1006            & 0.8334            & 1.4186            & 0.9609            & 1.1359            & 1.0899            \\ \midrule
Ours            & \textbf{0.2885}   & \textbf{0.2653}   & \textbf{0.4127}   & \textbf{0.2708}   & \textbf{0.2735}   & \textbf{0.3022}   \\ \bottomrule
\end{tabular}
}
\label{tab:quant_short_consis}
\vspace{-3mm}
\end{table}

We present the long-range consistency score in Table ~\ref{tab:quant_long_consis}.
Specifically, we use every two far-away views, i.e., the $t-th$ and $(t-7)-th$ frames in the testing videos, to compute the consistency score.
Since the distance between two views is larger, the consistency scores of all methods in this experiment are higher than those in the short-range study.
Although the video stylization baselines generally better preserve the short-range consistency than the image stylization ones, they fail to maintain the consistency between two far-away views due to the error accumulation.
In contrast, the proposed method is capable of synthesizing images that are both short-range and long-range consistent.

\begin{table}[!h]
\centering
\renewcommand{\arraystretch}{1.1}
\setlength{\tabcolsep}{5pt}
\caption{\textbf{Quantitative comparisons on long-range consistency.}
We compute the consistency score (the lower the better) between stylized images at two far-away novel views.
}
\resizebox{\columnwidth}{!}{
\begin{tabular}{l|ccccc|c}
\toprule
Methods         & Family            & Francis           & Horse             & Playground        & Truck             & Average           \\ \midrule
AdaIN           & 6.4704            & 5.5091            & 5.5914            & 6.4771            & 5.2145            & 5.8526            \\
WCT             & 6.7233            & 6.2752            & 6.8781            & 6.3403            & 6.6640            & 6.5767            \\
LST             & 8.0778            & 6.1186            & 8.1056            & 8.6210            & 9.3647            & 8.0575            \\
TPFR            & 4.2423            & \textbf{2.5301}   & 5.0199            & 5.1047            & 3.1312            & 4.0056            \\ \midrule
ReReVST         & 4.8321            & 4.5702            & 4.3904            & 5.8077            & 4.4881            & 4.8177            \\ 

MCCNet          & 5.7786            & 4.1305            & 4.6071            & 5.3677            & 4.7280            & 4.9224            \\ \midrule
Ours            & \textbf{3.5101}   & 2.8598            & \textbf{2.5637}   & \textbf{3.2701}   & \textbf{2.7423}    & \textbf{2.9892}  \\ \bottomrule
\end{tabular}
}
\label{tab:quant_long_consis}
\vspace{-3mm}
\end{table}

\subsection{Limitations}
\label{sec:discussion}
\begin{figure}[!h]
    \centering
    \includegraphics[width=\columnwidth]{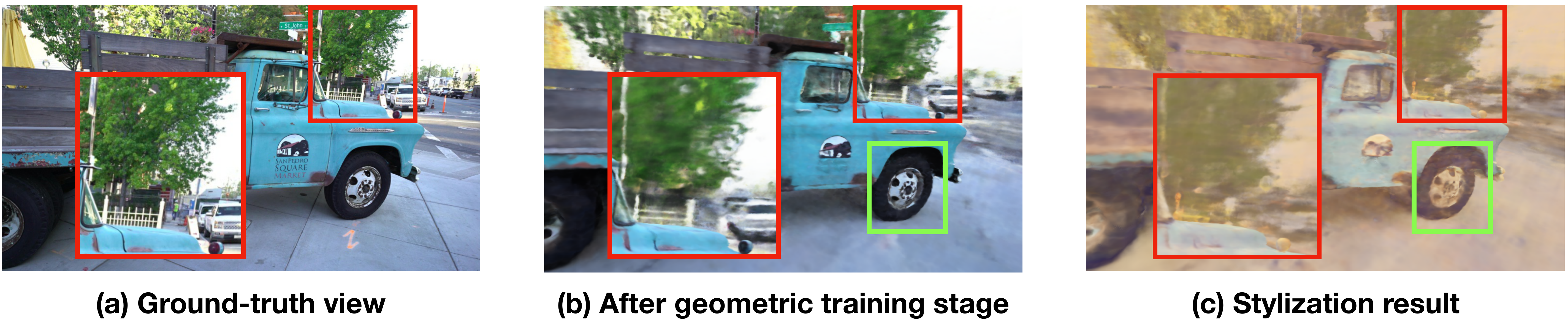}
    \caption{\textbf{Failure cases.}
    \students{Our proposed method is limited by the backbone NeRF model. The red boxes show a negative example that the details are blurry while the green boxes show a positive example that the details are clear.}
    }
    \label{fig:limitation}
\end{figure}

The quality of the stylization results is limited by the backbone NeRF model.
As the red boxes shown in Figure~\ref{fig:limitation}, the proposed method produces blurry stylization results since the backbone model fails to capture the details of the trees.
In contrast, the details of both the original and stylized wheels demonstrated in the green boxes are clear.

\vspace{-3mm}

\section{Conclusions}
In this paper, we propose a NeRF model for transferring arbitrary styles to complex 3D scenes.
We design a hypernetwork to predict the appearance-related parameters in the NeRF model to stylize the 3D scene according to the input reference (style) image.
In addition, we develop a two-stage training strategy along with the patch sub-sampling algorithm to learn the hypernetwork.
Qualitative and quantitative results validate that the proposed method renders high-quality novel view images with the desired style.

\noindent\textbf{Acknowledgement.}~This project is supported by MediaTek Inc. and MOST 110-2636-E-009-001. We are grateful to the National Center for High-performance Computing for computer time and facilities.

{\small
\bibliographystyle{ieee_fullname}
\bibliography{egbib}
}

\end{document}